\pdfoutput=1
\documentclass[%
 preprint,
 amsmath,amssymb,
 aps,
prl
]{revtex4-2}

\usepackage[english]{babel}
\usepackage{graphicx}
\usepackage{dcolumn}
\usepackage{bm}
\usepackage{nicefrac}
\usepackage{hyperref}

\usepackage{soul}
\usepackage[usernames,dvipsnames]{xcolor}
\usepackage{tikz}
\usepackage{comment}

\usepackage[protrusion=false,expansion=true,kerning=true,babel=true,final]{microtype}

\graphicspath{{images/}}

\begin{document}

\preprint{APS/123-QED}

\title{Design of bistable soft deployable structures via a Kirigami-inspired
planar fabrication approach}


\author{Mrunmayi Mungekar$^{1}$}
\author{Leixin Ma$^{1}$}\email{L.M.: leixinma@g.ucla.edu}
\author{Wenzhong Yan$^{1}$}
\author{Vishal Kackar$^{1}$}
\author{Shyan Shokrazadeh$^{1}$}
\author{M. K. Jawed$^{1}$}\email{M.K.J.: khalidjm@seas.ucla.edu}

\affiliation{\footnotesize 
$^1$University of California, Los Angeles, Department of Mechanical and Aerospace Engineering, 420 Westwood Plaza, Los Angeles, CA, USA 90024
}

\begin{abstract}
Fully soft bistable mechanisms have shown extensive applications ranging from soft robotics, wearable devices, and medical tools, to energy harvesting. However, the lack of design and fabrication methods that are easy and potentially scalable limits their further adoption into mainstream applications. Here a top-down planar approach is presented by introducing Kirigami-inspired engineering combined with a pre-stretching process. Using this method, Kirigami-Pre-stretched Substrate-Kirigami trilayered precursors are created in a planar manner; upon release, the strain mismatch---due to the pre-stretching of substrate---between layers would induce an out-of-plane buckling to achieve targeted three dimensional (3D) bistable structures. By combining experimental characterization, analytical modeling, and finite element simulation, the effect of the pattern size of Kirigami layers and pre-stretching on the geometry and stability of resulting 3D composites is explored. In addition, methods to realize soft bistable structures with arbitrary shapes and soft composites with multistable configurations are investigated, which could encourage further applications. Our method is demonstrated by using bistable soft Kirigami composites to construct two soft machines: (i) a bistable soft gripper that can gently grasp delicate objects with different shapes and sizes and (ii) a flytrap-inspired robot that can autonomously detect and capture objects.
\end{abstract}

\pacs{Valid PACS appear here}

\maketitle

\section{Introduction}

The rapid development of soft robotics has recently attracted increasing attention from biology, chemistry, materials science, and engineering~\cite{rus2015design}. Fully soft machines and robots have shown the capability of outperforming conventional rigid counterparts in terms of adaptability, robustness, and safety~\cite{whitesides2018soft}. Recently, bistable mechanisms---possess two stable equilibrium states---have been introduced into soft robotics for achieving high-performance and multi-functionalities~\cite{chi2022bistable}. They have demonstrated various innovative applications in fast grasping~\cite{baumgartner2020lesson,lerner2020design,sun2020tuning}, shape reconfiguration~\cite{chen2022spatiotemporally,faber2020dome}, information storage~\cite{chen2021reprogrammable,yasuda2017origami}, high-speed locomotion on ground~\cite{gorissen2020inflatable,kim2021autonomous,xiong2022fast} and water~\cite{chi2022snapping,chen2018harnessing}, adaptive sensing~\cite{riley2022neuromorphic}, and mechanical oscillation~\cite{rothemund2018soft,yan2021cut}, computation~\cite{preston2019digital,yan2021origami}, and feedback controls~\cite{drotman2021electronics} towards achieving autonomy for fully soft machines. Fabrication is non-trivial for fully soft bistable mechanisms, including both pre-shaped and pre-buckled ones~\cite{pal2021exploiting}. Pre-shaped bistable mechanisms are either accomplished with vacuumed-involved casting procedures with customized molds~\cite{liu2022soft,wang2020soft} or with free-form fabrication techniques (i.e., three dimensional printing)~\cite{yap2016high} with limited scalability and long fabrication times~\cite{jones2021bubble}. Likewise, pre-buckled ones often require constrained boundary to precisely impose buckling load~\cite{thuruthel2020bistable,wu2021fast}, which could increase the fabrication difficulty and restrain their applications. Therefore, it is desired to introduce a method for fabrication and programming of fully soft bistable mechanisms that are easy to generate and potentially scalable, which could encourage their further adoption into mainstream applications.

Recently, pre-stretching method on generating three dimensional (3D) structures and mechanisms from planer bilayered precursors (structures before 3D deployment) has drawn increasing attentions thanks to their scalability and ease for fabrication~\cite{xu2015assembly, fan2020inverse, liu2020tapered}. The planer precursors often consist of a pre-stretched substrate with a bonded top layer; upon released, the substrate imposes compressive load on the top layer to force the bilayered precursor into 3D morphology~\cite{yan2016controlled}. However, to introduce bistability or even multistability, a spatial variation of thickness and selected bonding pattern for the top layer are necessary~\cite{luo2019mechanics}. These two technologies could enable the creation of more delicate 3D configurations while sacrificing the ease of fabrication. 

A promising alternative method is to bring in Kirigami-inspired engineering onto the top layer (with uniform thickness); the designated pattern on the top layer could potentially guide the resulting composite to deform into a pre-programmed geometry without requiring neither a spatial variation of thickness of the top layer nor specified bonding pattern~\cite{cui2018origami,zhang2022shape,miha2022soft, ma2022soft}. This Kirigami-inspired pre-stretching method has enabled various applications, including deployable 3D shapes~\cite{cui2018origami}, self-assembly structures~\cite{van2020Kirigami}, etc.~\cite{zhang2017printing}. 
Yet, the capability of introducing bistability within this design paradigm has not been demonstrated. Specifically, a scalable and simple manufacturing method that enables the usage of the uniform material and bonding could create a pathway for broad adoption of fully soft bistable mechanisms into intelligent soft machines and robots.

\begin{figure*}[!ht]
\centering
\includegraphics[trim=0in 9.6cm 0in 0cm, clip=true, width=6.7in]{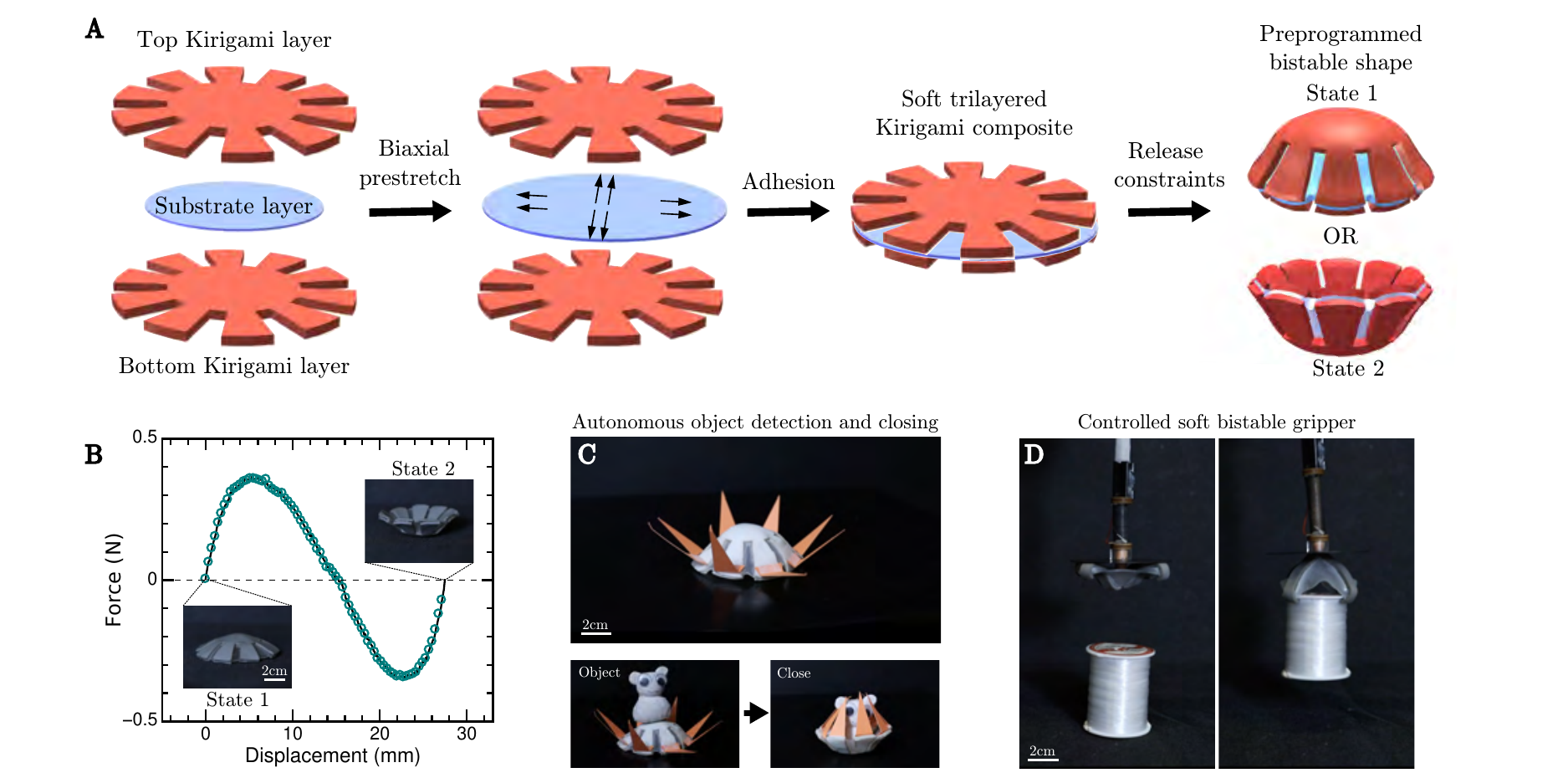}
\caption{Overview of the bistable soft structure design. (A). The combination of two Kirigami layers and strain mismatch creates 3D soft structures that are fully soft and bistable. (B) The force-displacement curve demonstrates that the trilayer soft structure can snap-through from one stable configuration to the other. (C) The bistable structures can be applied to create a flytrap-inspired robot allowed for autonomous sensing and fast actuation to capture objects. (D) A bistable pyramid can function as a gripper to grasp delicate objects with different size and shape.
}
\label{fig:overview}
\end{figure*}

Here we propose an easy and scalable process to create fully soft bistable mechanisms from 2D preprogrammed precursors. This approach introduces the bistability through bonding one central pre-stretched substrate with two identical Kirigami top and bottom layers. Once released, the trilayered composite deforms into a 3D free-buckling configuration with two stable equilibrium, as schematically shown in Figure~\ref{fig:overview}{\em A} and {\em B}. Our method allows us to only use commercially available soft elastomer sheet materials, which could vastly simplify the fabrication processes. We combine physical experiment, finite element simulation, and composite shell theory to explore and predict the transition from 2D to 3D shapes and the properties of the resulting bistable mechanisms. We show methods to create soft bistable structures with arbitrary shapes and sizes by combining optimal design results from machine learning algorithm with scaling analysis. In addition, we explore the multistability of proposed trilayered composites through experiment and simulation. We demonstrate this generic design and fabrication method by creating (i) an universal, nondestructive soft gripper that allows stable grasping of delicate objects such as a strawberry or a butter packet; and (ii) a flytrap-inspired robot that can autonomously sense and swiftly capture objects (see Figure~\ref{fig:overview}{\em C} and {\em D}, and Movie S1).  This work opens a new avenue to generate fully soft bistable mechanism with potential applications in wearable devices, soft robotics, and multifunctional medical devices.

\section{Results and Discussion}
\subsection{Design concept}

Figure ~\ref{fig:overview} presents our concept of designing bistable soft composites. We start with a substrate layer and two identical outer layers with the same Kirigami design cutouts (Figure~\ref{fig:overview}{\em A}). The Kirigami layers in this example resemble a lotus shape. 
 First, a biaxial stretch is induced on the substrate layer. 
Then, the two unstretched Kirigami layers are aligned and bonded at the top and the bottom of the pre-stretched substrate layer, generating symmetry to the composite about the substrate (Figure~\ref{fig:overview}). The material removal in the Kirigami layers can help the laminated trilayered structure to achieve targeted global shapes with smooth transition instead of wrinkling in the local regions~\cite{paulsen2016curvature}.
Once released, the strain mismatch between the three layers induces out-of-plane buckling since bending is less energetically expensive than compression for thin shells~\cite{pezzulla2016geometry}.   



\subsection{Desktop-Scale Physical Experiments}

\begin{figure*}[!ht]
\includegraphics[trim=0in 4.2cm 0in 0cm, clip=true, width=6.7in]{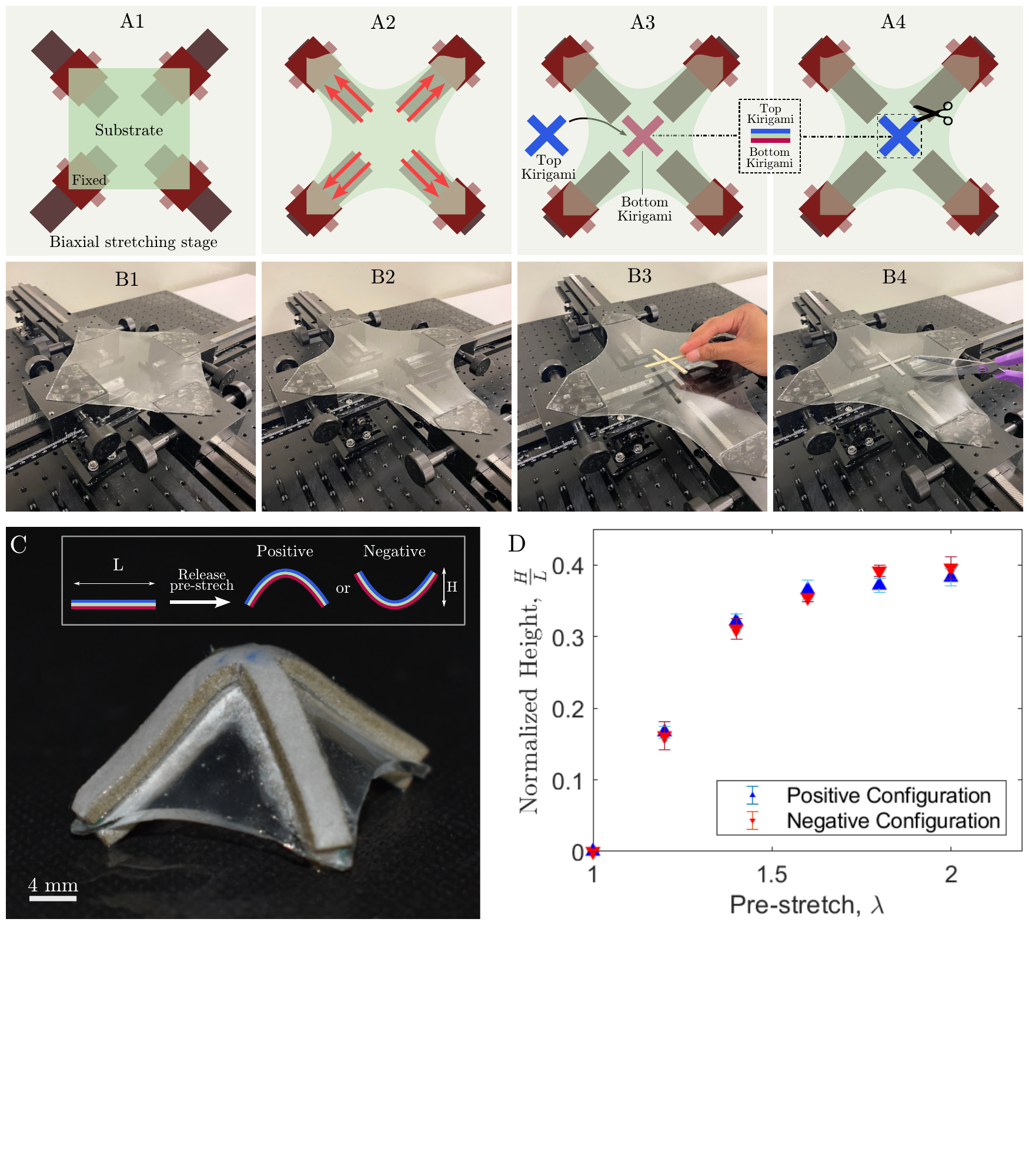}
\centering
\caption{Experimental setup. (A1-A4) CAD model of the experimental setup (B1) Schematic representation of the system consist of two-knob stages (1)  on 250 mm tracks (2) (B2) Snapshot of the system when the substrate layer (3) is stretched (B3) Attach one Kirigami layer (4) below the substrate layer, and the other one (5) on top of the substrate layer (B4) Release the pre-stretch by cutting along the outline of the circular substrate.(C) Side view of the trilayer composite structure, and the free buckling shape of the trilayer composite structure (D) Variation of normalzied height as a function of the pre-stretch ($L = 60$ mm)}
\label{fig:fig2}
\end{figure*}

Figure~\ref{fig:fig2} presents the experimental setup for fabricating trilayered bistable structures. 
The experimental setup consists of four linear translation stages (250 mm travel, Thorlabs, Inc.). The substrate and the Kirigami layers are made of hyper-elastic materials. The material properties are listed in Table \ref{table:matparam}. We fit the experimentally measured stress-strain curve using the Mooney Rivlin material model. The material constants for the Kirigami layer are $C_1^{kg}$ and $C_2^{kg}$, while the constants for the substrate layer are denoted as $C_1^{s}$ and $C_2^{s}$. An arbitrarily large substrate layer is fixed with its four corners attached onto the four stages, respectively (Figure~\ref{fig:fig2}{\em A}1 and {\em B}1). Then, to impose the biaxial stretch upon the substrate, we induce the substrate layer with the same amount of stretch in both directions (Figure~\ref{fig:fig2}{\em A}2 and {\em B}2). 
Once stretching is complete, two Kirigami layers are attached on the top and bottom of the substrate layer directly since both sides of layers already have strong adhesive agent (Figure~\ref{fig:fig2}{\em A}3 and {\em B}3). The excess substrate is cut away along a defined bonding box, which is designed according to the desired resulting shape (Figure~\ref{fig:fig2}{\em A}4 and {\em B}4). Once released, this planar composite structure then morphs to either one of the two possible stable configurations arising due to the strain mismatch within layers. 


For our experimental study, we chose cross-shaped Kirigami layers and square-shaped substrate layers. 
With the appropriate value of pre-stretch $\lambda$, these geometric designs of the layers provide a deformed shape that looks like a pyramid. Figure~\ref{fig:fig2}{\em C} displays one such soft pyramid bistable structure. The length of the 2D precursor is denoted as $L$ while the height of the buckled 3D pyramid is $H$ as shown in the insert of Figure~\ref{fig:fig2}{\em C}. We define the configuration it immediately converged to after removing the boundary condition as the “positive” configuration, while the inverted one as the “negative” configuration. Before the measurements were performed, both ``positive" and ``negative" configurations were given a sufficient period of time to settle down. As the pre-stretch increases, the normalized height (= $H/L$) of the pyramid increases in a nonlinear manner, as shown in Figure~\ref{fig:fig2}{\em D}. The normalized height for the ``positive" and ``negative" configurations were close to each other for this size of composite structure with $L = 60$ mm.

\subsection{Finite element-based numerical simulations}
The finite element software, Abaqus~\cite{abq}, was used to model the nonlinear large deformation of the hyper-elastic composite structures. In the simulations, we used the same pyramid shape and simulated the hyper-elastic materials of the substrate and the two Kirigami layers using Mooney Rivlin material models. Solid elements (C3D6)  were used to model the layers of the composite structure. Each layer of the composite was modelled with multiple sub-layers of elements to accurately predict the deformation of the structure along its thickness. Each sub-layer was meshed with two sets of 4521 nodes, with the substrate and the kirigami having 2198 and 6046-node prismatic elements per sub-layer, respectively. The symmetry of the Kirigami was taken into account for the mesh. Furthermore, the elements at the interface of the material layers shared nodes constraining their deformation in a way similar to the layers being bonded together. 
The center of the composite structure was set fixed. The pre-stretch was induced by describing an initial stress condition on the substrate layer. This initial stress was calculated as the Cauchy stress ($\sigma$) arising in a Mooney-Rivlin hyperelastic material (Parameters: $C\textsubscript{1}$, $C\textsubscript{2})$ due to a given equibiaxial stretch ($\lambda$).
\begin{equation}
    \begin{aligned}
         \sigma = 2C_{1}(\lambda^{2}-\frac{1}{\lambda^4}) - 2C_{2}(\frac{1}{\lambda^2}-\lambda^{4})
    \end{aligned}
\end{equation}
A nonlinear quasi-static deformation process was simulated with a very small but non-zero distributed force on the structure to force it into one of the two possible stable configurations, which we termed as the ``positive" configuration.

One numerical simulation took about 1-3 minutes wall clock time on a desktop computer (Ryzen 2950wx CPU @ 2.4 GHz). To understand the variation of the final shape with pre-stretch, the simulation was run for multiple values of pre-stretches. The height of the structure was studied by procuring the maximum and minimum values of the out-of-plane nodal coordinates. Further details on this study have been described in Section \ref{sec:effect}.

We also study the other stable configuration than the one the simulation initially provides. This is simulated by modelling a force-based actuation on the obtained configuration. Force is applied on four symmetrically placed nodes on the arms of the cross-shaped Kirigami layer, pushing the structure into the other stable configuration. This actuation provides us with details about the ``negative" configuration of the structure as well as brings to light the multiple stable modes that a structure could have as further described in Section \ref{sec:multistable}.

\subsection{Analytical derivation}

\begin{figure*}[ht]
\centering
\includegraphics[width=\textwidth,keepaspectratio]{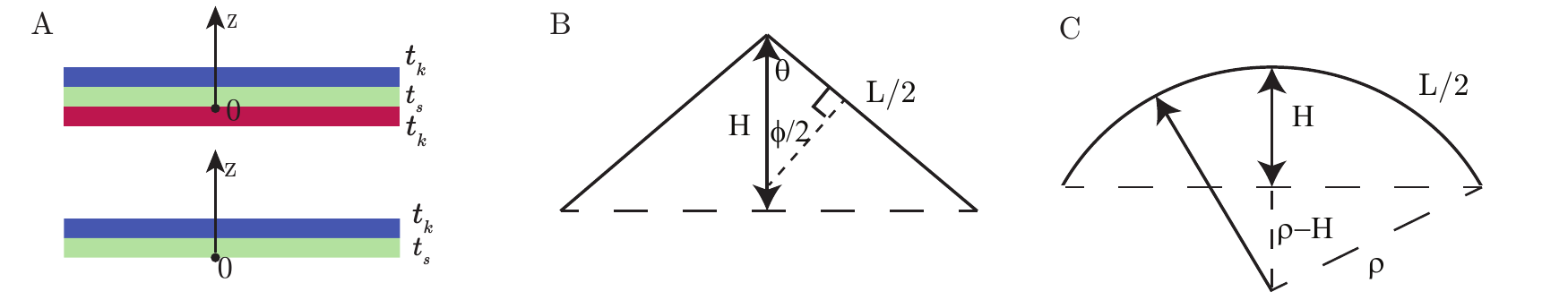}
\caption{(A) Cross-sectional view of the trilayer and bilayer composite structure (B) Sketch of the side view for a pyramid with height $H$. (C) Sketch of the side view for a spherical cap with height $H$. }
\label{fig:sketch}
\end{figure*}
Figure~\ref{fig:sketch}{\em A} shows the cross-sectional views of the soft Kirigami composite structure before bulking. The thickness of the substrate and Kirigami layers are denoted as $t_s$ and $t_k$, respectively. We assume that after the free buckling, the initial stretching energy of the substrate layer is transformed into the bending energy of the whole composite structure. This is because that bending energy is less energetically expensive than compression for thin shells~\cite{pezzulla2016geometry}.
Considering the hyperelastic property of the system, the stretching-induced energy is,
\begin{equation}
    \begin{aligned}
         \textbf{E}_s &= [C_1(2\lambda^2+1/\lambda^4-3)+C_2(\lambda^4+2/\lambda^2-3)] At_s
    \end{aligned}
\end{equation}
where $\lambda= L_f/L=1+\epsilon$ is the pre-stretch, which is the ratio of the stretched substrate size $L_f$ over the unstretched size $L$. $A$ is the area of the unstretched substrate. For the square-like substrate, $A= (L/\sqrt{2} \lambda)^2$.

For small strain, we can approximate the stretching-induced energy for the substrate layer as,
\begin{equation}
    \begin{aligned}
         \textbf{E}_s=\frac{1}{2}AC_s\epsilon^2
    \end{aligned}
\end{equation}
where $C_s= E_st_s/(1-\nu_s^2)$. $E_s$ and $\nu$ are the Young's modulus and Poisson's ratio of the substrate layer, respectively. 

Meanwhile, the bending energy can be calculated as,

\begin{equation}
    \begin{aligned}
         \textbf{E}_b 
        &= \frac{1}{2}AD_\textrm{eq}\kappa^2
    \end{aligned}
\end{equation}

\noindent where $D_\textrm{eq}$ is the equivalent bending stiffness of the trilayer, which is derived in the supplementary material. 
From the observation of the deformed 3D shapes, we further assume that the deformed shape is close to a pyramid shape, with the side view sketched in Figure~\ref{fig:sketch}{\em B}. For the pyramid like shape, the curvature is related by the angle $\phi$, the half-length of the cross shape $L/2$, and the height of the pyramid $H$ via the equations as follows,
\begin{equation}
    \begin{aligned}
        \kappa= 4\tan(\phi/2)/L 
        &=         \sqrt{\frac{C_s}{D_\textrm{eq}}}{\frac{\epsilon}{1+{\epsilon}}}
    \end{aligned}
    \label{part1}
\end{equation}

\begin{equation}
    \begin{aligned}
        H= L \sin(\phi/2))/2
    \end{aligned}
    \label{part2}
\end{equation}

Hence, equalizing the bending energy and stretching energy, and combining equation \ref{part1} and \ref{part2} gives,
\begin{equation}
    \begin{aligned}
        \frac{H}{L}= \sin(\phi/2)
        =\sin[\arctan{(\kappa L/4)}]/2
    \end{aligned}
    \label{eq50}
\end{equation}
Since $\arctan{(\kappa L/4)}$ function is bounded between 0 and $\pi/2$, the normalized height, $H/L$, is bounded between 0 and 0.5. This helps explain why in both simulations and experiments, the normalized heights plateau at very large amount of pre-stretches in in Figure \ref{fig3}.
It can be found that the simplified analytical model agrees well with the experiments and simulations, which further verifies the concept of shape-forming trilayer structures.

For other 3D shapes close to spherical caps in Figure~\ref{fig:sketch} C, the relationship between normalized height and curvature are found to be presented in Equation~\ref{rel}. The detailed derivations are in the supplementary material. 
\begin{equation}
    \begin{aligned}
        \frac{H}{L}= \frac{\kappa L}{8}
    \end{aligned}
    \label{rel}
\end{equation}
\subsection{Effect of size and pre-stretch on free buckling shapes}
\label{sec:effect}

In this section, we focus on the cross-shaped Kirigami layers with square-shaped substrate 
as shown in Figure~\ref{fig:fig2}. The thickness of the material is denoted as $t$ (including $t_s$ and $t_k$ for the substrate and Kirigami layer, respectively), the length and width of the kirigami are denoted as $L$ and $w$, respectively.

\begin{figure*}[!ht]
\includegraphics[width=\textwidth,keepaspectratio]{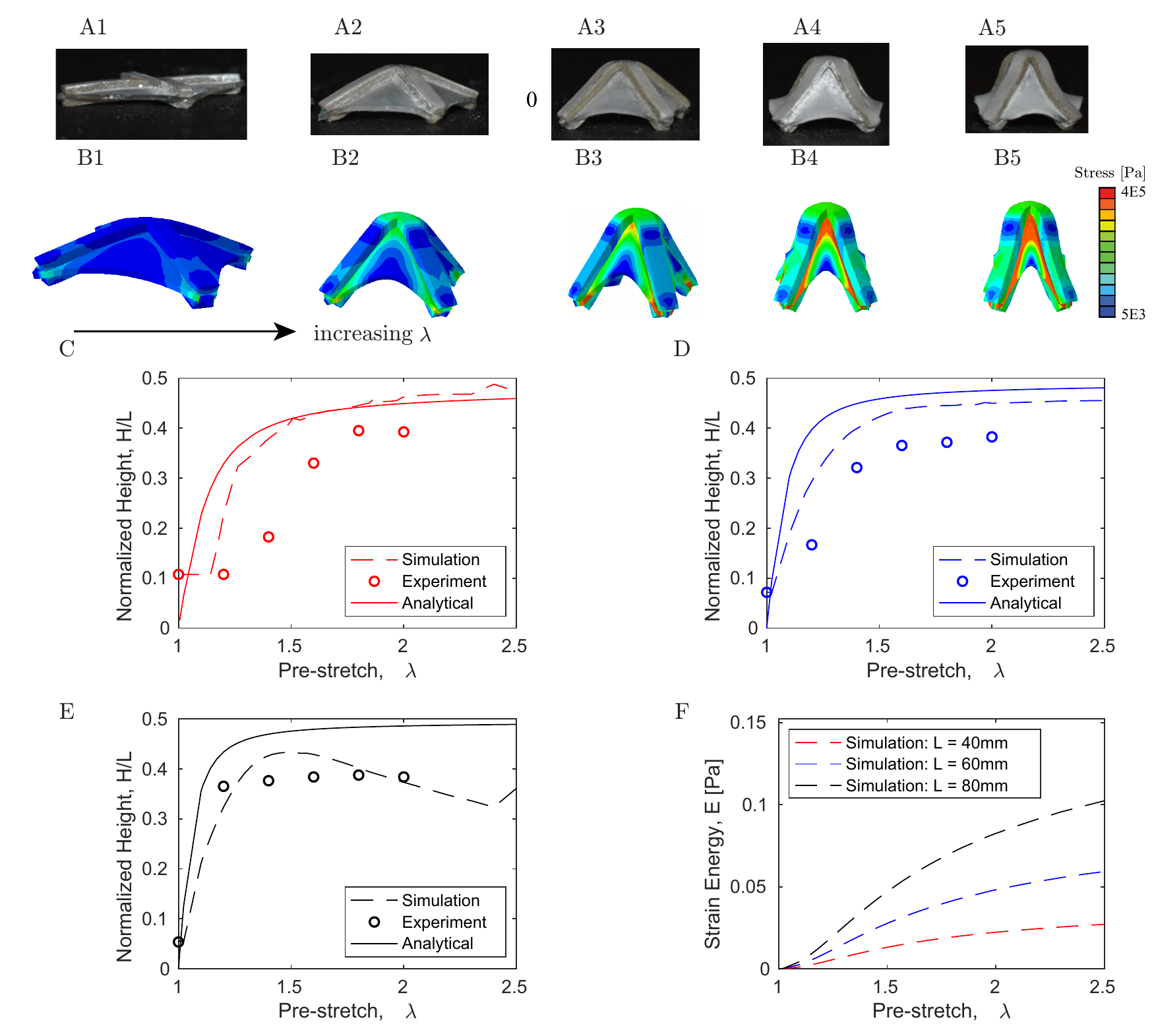}
\centering
\caption{
(A) Comparison of the free-buckling shapes in (A1-A4) simulation and (B1-B4) experiment for the cross-shaped Kirigami composite structure (L=40 mm) at different amount of pre-stretch $\lambda = 1.2, 1.4, 1.6, 1.8, 2.0 $. The colorbars correspond to the distribution of stress in the deformed trilayer structure (C-E) The variation of the normalized height of the free-buckling shape as a function of the pre-stretch. Three examples showing different length/thickness ratio with diameter/width ratio fixed. (C) $L/t_s = 4000$. (D) $L/t_s = 6000$. (E) $L/t_s = 8000$. (F) Variation of the energy landscape at different pre-stretch. Below a certain threshold, the composite structure hardly buckles.
}
\label{fig3}
\end{figure*}

For a fixed size of kirigami composite structure, we first investigate the effect of prestretch on the buckling shapes. A few examples of Kirigami composites with different amount of pre-stretches are presented in Figure~\ref{fig3} {\em A} and {\em B}. The geometry in simulations agrees well with the results of experiments. The height of the composite structure increases with the amount of pre-stretch, as shown in Figure~\ref{fig3}{\em C} to {\em E}. This is because with the increase in strain energy in Figure~\ref{fig3}{\em F}, the converted shell bending energy becomes larger. There exists a certain threshold, below which the composite structure hardly buckles. 
Despite the assumptions introduced in the analytical model, the experiment, simulation and analytical model exhibit similar trend. As pre-stretch increases, the maximum heights of these pyramid-like 3D shapes are bounded by a certain maximum.

We further increase the length $L$, but fix the length/width ratio, $L/w$, and the thickness of the composite structure $t$. This means that the ratio of length/thickness, $L/t$, is increased. Figure~\ref{fig3}{\em C}, {\em D} and {\em E} present the variation of normalized height at different $L/t$. At the same level of pre-stretch, the maximum normalized height for the pyramid tends to be larger for thinner composite structures (with larger $L/t$). Such a behavior can be understood from the scaling relationship in Equation \ref{eq50}.

\begin{figure*}[ht]
\includegraphics[trim=0in 3.6cm 0in 0cm, clip=true, width=1\textwidth]{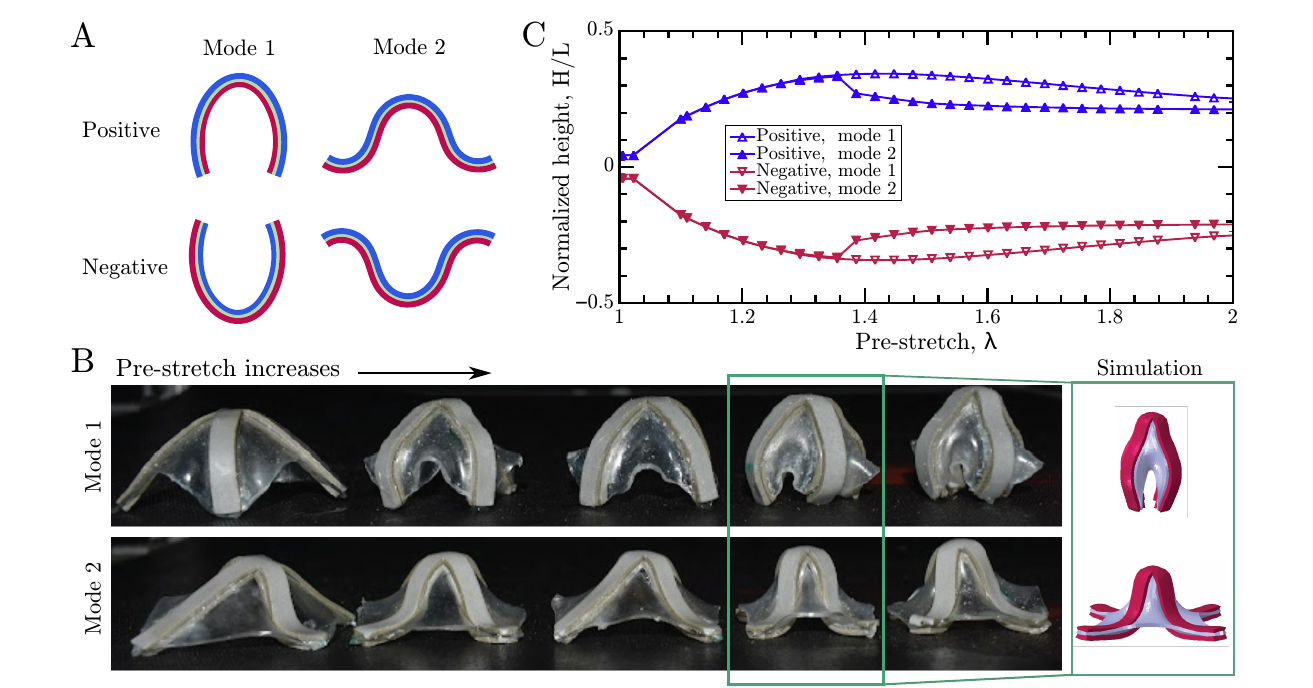}
\centering
\caption{
(A) Sketches of free-buckling mode shape 1 and mode shape 2 in the ``positive" and ``negative" configurations. (B) The 
free buckling shapes for the for the cross-shaped Kirigami composite structure ($L= 80$ mm) at different amount of pre-stretch $\lambda = 1.2, 1.4, 1.6, 1.8, 2.0 $. The Mode 1 and Mode 2 are also predicted in simulations. (C) The simulated variation of normalized height of the two modes in the ``positive" and ``negative" configurations as a function of pre-stretch $\lambda$.
}
\label{figmult}
\end{figure*}

\begin{figure*}[ht]
\includegraphics[trim=0in 0cm 0in 0cm, clip=true, width=6.9in]{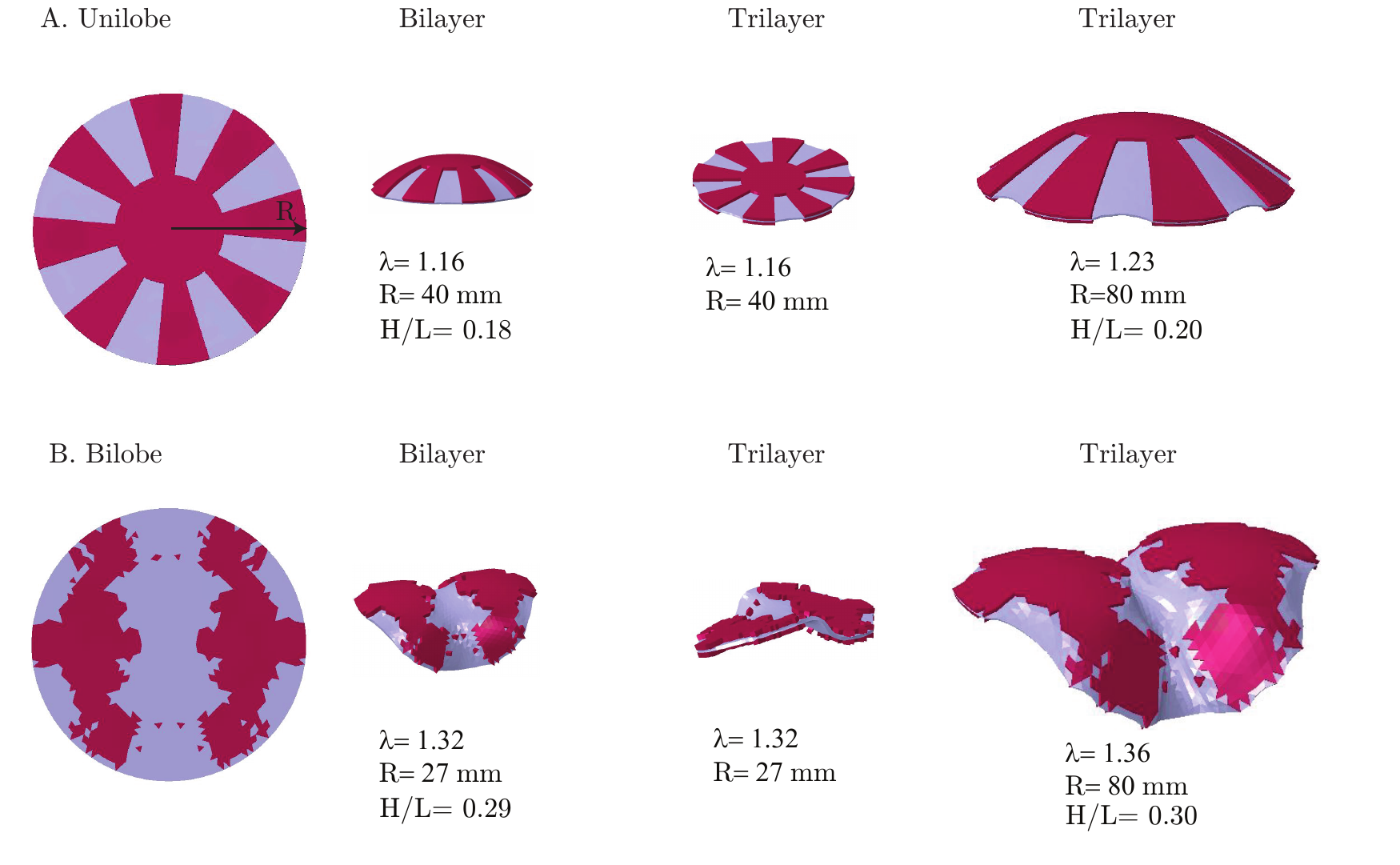}
\centering
\caption{Comparison of the free buckling shape for bilayer and trilayer soft Kirigami composite. The Kirigami patterns is fixed, and the structures can create two target shapes (A) Unilobe structure. (B) Bilobe structure. First column: Kirigami patterns in red and circular substrate in blue. Second column: Bilayer structure with specific combination of pre-stretch and initial radius. Third column: Trilayer structure with the same combination of pre-stretch and initial radius as the bilayer counterpart. Fourth column: Trilayer structure with adjusted pre-stretch and radius. 
}
\label{fig4}
\end{figure*}

\subsection{Multistability}
\label{sec:multistable}

As $L$ increases to $80$ mm, we find there could be multiple free-buckling mode shapes in the positive and negative configurations, as shown in Figure~\ref{figmult} {\em A} and {\em B}. 
This behavior is not observed for thicker composite structures with larger $L/t$ in Figure~\ref{fig3}. For the kirigami composite of $L= 80$ mm, a systematic variation finds that, at a small pre-stretch, the free buckling shape is dominated by the mode 1 defined in Figure~\ref{figmult}{\em A}. As the amount of pre-stretch increases, the mode 2 type emerges. The four edges of the cross-sections start to buckle locally. In the two modes, the cross shapes buckle in different directions. This leads to the difference in the height of the two modes presented in Figure~\ref{figmult}{\em C}. Considering the symmetry of the trilayered structure, the soft structures in this case have four stable configurations. Such a multistable behavior further functionalizes the structure to possess more mechanical properties. This also opens the door of conducting systematic analytical research to understand the mechanism behind the multi-stability phenomena in the future. 

\subsection{Creating the trilayer bistable structures that achieve target 3D shape}

Based on the same design concept of combining Kirigami structures and strain mismatch, Miha et al. and Ma et al.~\cite{miha2022soft} for the first time demonstrated that the planar bilayered composite can be deformed to 3D soft structures. Ma et al.~\cite{ma2022soft} further developed a machine learning-aided approach to explore the optimal Kirigami cuts, pre-stretch and Kirigami size that achieves targeted 3D shapes, such as peanuts and flowers. The previous research enables rapid prototyping of soft Kirigami structures through a Kirigami bilayered composite structure.
If we continue to add more Kirigami layers in the system, we can introduce bistability for these soft kirigami structures, which can be used for wider applications, such as biomimetics ~\cite{cui2018origami} and soft actuators ~\cite{liu2022soft}. 

Even though the proposed machine learning approach by Ma et al. ~\cite{ma2022soft} reduces the number of searches from millions of designs to order of hundreds, it still takes about 6 hours of computational time on a desktop computer (Ryzen 2950wx CPU @ 2.4 GHz). In this section, we want to test whether we can achieve similar targeted 3D bistable structures just by modifiying the size and the amount of pre-stretch while keeping the optimized Kirigami patterns the same as the bilayered counterparts. To achieve this goal, we make use of the scaling analysis derived from the energy conservation.

The 3D shapes we focus on in this section are the unilobe shape and the bilobe shape (Figure~\ref{fig4}). For these 3D shapes, we can assume the curvature is approximately the same everywhere. In this case, the relationship between the curvature and height is derived in the supplementary information. Then, equalizing the bending energy and stretching energy, the normalized height $H/L$ can be approximated by 
\begin{equation}
    \begin{aligned}
        \frac{H}{L}= \frac{L}{8}\sqrt{\frac{C_s}{D_\textrm{eq}}}{\frac{\epsilon}{1+{\epsilon}}}
    \end{aligned}
    \label{eq5}
\end{equation}

For the structure of a certain radius $R$, the bending stiffness ${D_\textrm{eq}}$ for the trilayer is about 3 times of the bilayer counterpart. 
Figure~\ref{fig4} shows two examples where the Kirigami patterns in the first column can be used to create a target unilobe and a bilobe bilayer structures in the second column ~\cite{miha2022soft,  ma2022soft}. The material properties for the substrate and Kirigami layers are shown in Table~\ref{table:matparam}. The material constants for the Kirigami layer leading to unilobe structures are  $C_1^{kg}$ and $C_2^{kg}$. While the constants for kirigami that creates the bilobe shape are $C_1^{kw}$ and $C_2^{kw}$. If we apply the same amount of pre-stretch to the trilayer structure of the same size as the bilayer counterpart, then the free-buckling shape is still flat in the plane. This is due to the increased stiffness in the system. 
To create the targeted 3D shapes of similar height ratio $H/L$, the pre-stretch and size need to be adjusted according to Equation~\ref{eq5}. The last column in Figure~\ref{fig4} shows some examples when the pre-stretch varies near the value for the bilayer counterpart but the radius of the structure is increased, we can still fabricate unilobe and bilobe shapes of similar $H/L$ using these trilayer composite structures. 

There also exists a limiting condition when the scaling relationship can be further simplified. When the thickness of the substrate is very thin, and the bending stiffness of the substrate is negligible, the equivalent bending stiffness of the trilayer is about 8 times of the bilayer counterpart with the same size. Hence the ${\frac{\epsilon}{1+{\epsilon}}}$ or size $L$ needs to be increased to approximately $2\sqrt{2}$ of the bilayer structure.

\subsection{Applications of bistable, soft Kirigami composites}

\begin{figure*}[!ht]
\includegraphics[trim=0.4in 12cm 0.5in 3.5cm, clip=true, width=6.7in]{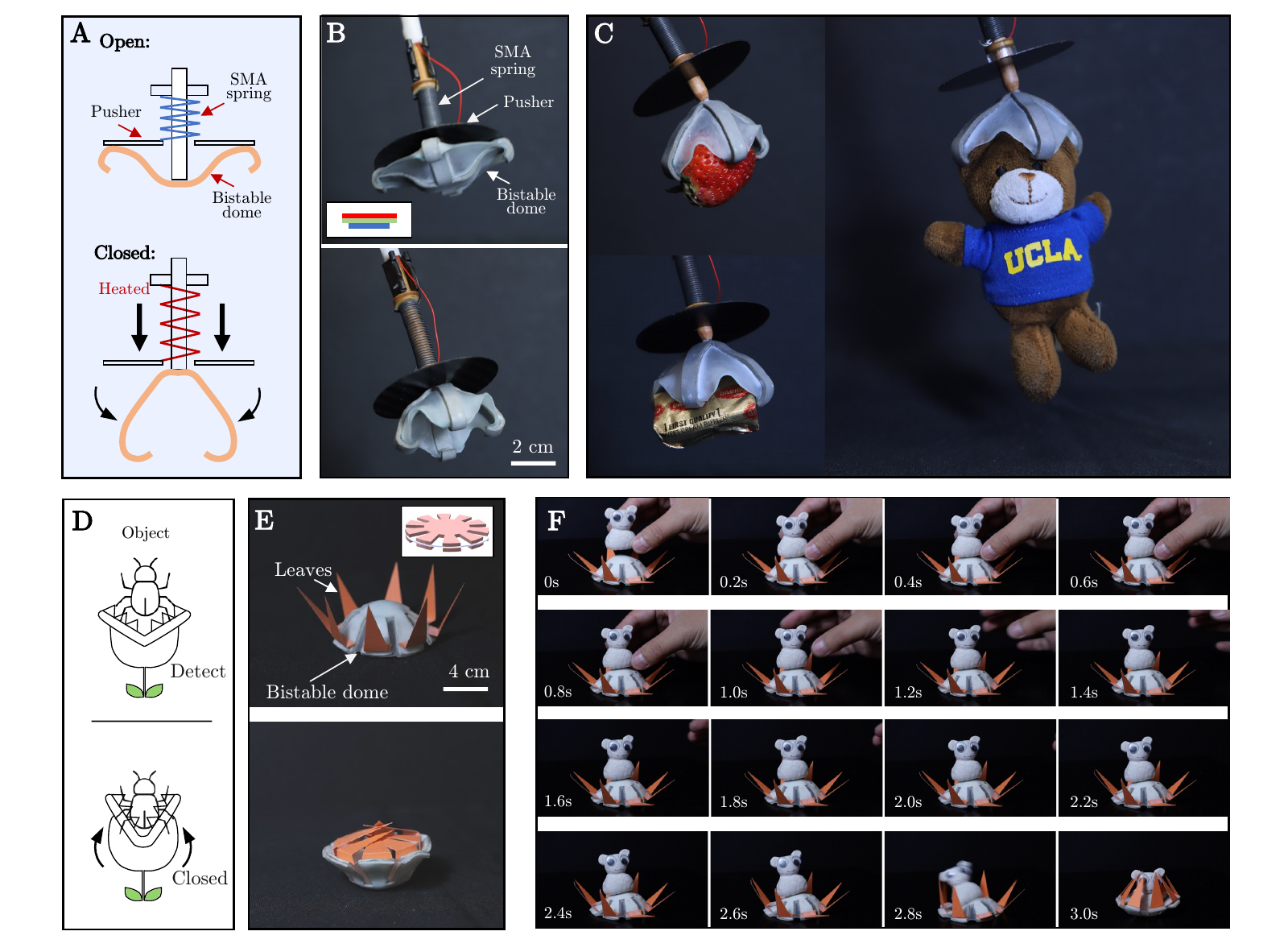}
\centering
\caption{Demonstrating bistable soft composites with two applications. (A) The mechanism of a soft griper composed of a bistable dome actuated by an extending shape memory alloy(SMA) spring. (B) The detailed structure of the gripper with labels. (C) The gripper has the ability to gently grasp delicate objects of different shapes and sizes. (D) \textit{Venus flytrap} can detect preys and close its leaves quickly to capture preys. (E) This strategy can be mimicked by mechanically embedding sensing and actuation into structures by using a bistable dome. (F) Snapshots show the autonomous object detection and quick closing of leaves without requiring external sensing, control, and actuation. To help overcome the energy barrier of the bistable dome, we attached a permanent magnetic bead on the bottom of the object while placing another one underneath the platform.}
\label{fig:demo}
\end{figure*}

To show the potential applications, we used bistable soft Kirigami composites to realize two soft machines: (i) A bistable soft gripper that can grasp delicate objects with different shape and size (see Figure~\ref{fig:demo}{\em A}-{\em C}); (ii) A flytrap-inspired robot that can autonomously and passively sense and capture objects (see Figure~\ref{fig:demo}{\em D}-{\em F}).  

The soft gripper is mainly composed of a bistable dome ($4.2$ gram, with cross-shaped Kirigami layers), an extending shape memory alloy (SMA) spring, and a mechanical pusher (see Figure~\ref{fig:demo}{\em A}). The center of the bistable dome is attached on the end of a rail. One end of the SMA spring is fixed on the rail while the other is joined with the pusher. The spring can be activated by Joule heating through electrical power. Thus, the bistable dome stays at the upward stable equilibrium when unactuated; the pusher can glide along the rail when the SMA spring is powered and eventually causes the dome to snap-through to the other stable state. To facilitate the grasping, we modified the original Kirigami pattern by extending the top Kirigami and substrate layers on the edges so that the added parts of the dome can form hook-shaped structures locally as shown in Figure~\ref{fig:demo}(B). This shows the potential of our method to generate more sophisticated 3D structures to enable more functionalities through a monolithic planar process. We show the ability of our gripper by grasping various fragile objects, including a strawberry (15 gram), a butter packet (9.7 gram), and an UCLA plush bear (12.6 gram) as shown in Figure~\ref{fig:demo}{\em C}. We also demonstrate sturdy grasping of a cable spool (25.3 gram, Figure~\ref{fig:overview}{\em D}, see supplementary video for details) under moderate disturbance. It is worth noting that we only need energy during the transition phase while no energy is required to keep the gripper closed, which could save considerable amount of energy for applications where long grasping time is necessary. 

The flytrap-inspired robot consists of a bistable trilayered dome (with lobe-shaped Kirigami layers) with triangular leaves (see Figure~\ref{fig:demo}{\em E}). The transition of the robot from the open state to the closed one happens by applying a minimum loading force. This snap-through transition with the intrinsic energy barrier can function as embedded sensing and fast actuation to autonomously capture object. We demonstrate this by placing a self-made toy (made from Plasticine) on the center of the robot; the loading causes the bistable dome to snap through and thus quickly close leaves to trap the object. The closing happened within 0.4 s (see Figure~\ref{fig:demo} {\em F}, see supplementary video for details) while the toy maintains intact thanks to the softness of the trap. To facilitate the transition, we harnessed magnetic force by attaching a magnetic bead on the bottom of the toy with another magnetic bead underneath the robot. Although this is out of scope of this paper, we could lower the energy barrier by using thinner sheet materials or reduce pre-stretch to further release the requirement for magnetic beads. This embedded sensing and actuation into soft material could reduce the complexity and weight, and increase the adaptability of resulting devices; without requiring external electronic control and actuation could also make our robots applicable for cases where conventional electronics can not survive. 


\section{Concluding Remarks}
In this paper, we numerically and experimentally studied the influential parameters on how an initially planar substrate layer and two Kirigami layers can be used to fabricate bistable soft 3D structures. The structure size and applied pre-stretch are found to be very critical in affecting the geometry and the stablility of the free-buckling shapes. By introducing analytical analysis to the optimized kirigami patterns for bilayer counterparts, we found that by carefully adjusting the pre-stretch and structural size, we are able to directly extending the 3D shapes designed for bilayers to trilayer structures. This combined approach provides a rapid way of designing and fabricating soft, bistable composite structures of different targeted 3D shapes. The proposed fabrication techniques, systematic parametric analysis, and mathematical modelings can lay a foundation in applications in soft robotics, and wearable devices etc. We demonstrate the advantage of using these soft bistable structures in engineering tasks that requires fast grasping and stable manipulation for delicate objects, and autonomous object detection and fast object capturing. 



\section{Source code}
\label{sec:sourceCode}
The source code for Finite Element Simulation for all conducted simulation tests can be found at https://github.com/StructuresComp/bistable-kirigami.

\section{Supplementary}

\subsection {Relationship between the curvature and the maximum height for spherical cap shapes}
We further assume that the deformed shape is close to a spherical cap shape with maximum height $H$, and has approximately constant curvature everywhere, as shown in Figure~\ref{fig:sketch}{\em C}.
The length of the curve $L$ can be approximated as, 

\begin{equation}
    \begin{aligned}
            (L/2)^2= \rho^2-(\rho-H)^2+H^2
    \end{aligned}
\end{equation}
where $\rho$ is the radius of the hemisphere, which is related to the curvature $\kappa$ via $\kappa= 1/\rho$.
Hence, 
\begin{equation}
    \begin{aligned}
        \frac{H}{L}= \frac{L}{8}\kappa
    \end{aligned}
    \label{spherickappa}
\end{equation}

\subsection {Equivalent bending stiffness for composite shells}

Figure~\ref{fig:sketch}{\em A} shows the cross-sectional view of the bilayer and trilayer composite structures.
The neutral axis of the trilayer is in the center of the substrate. 
Hence, the equivalent bending stiffness is,
\begin{equation}
    \begin{aligned}
        D_\textrm{tri} = n\frac{2{E_{k}[{t_k^3/12+t_k(t_k/2+t_s/2)^2}]+E_{s}[{t_s^3/12}]}}{{\left( {1 - {\nu ^2}} \right)}}+(1-n)\frac{E_{s}{t_s^3/12}}{{\left( {1 - {\nu ^2}} \right)}}
    \end{aligned}
    \label{be1}
\end{equation}
where $n$ is the fraction of the area covered by the kirigami layer. 
$t_s$ and $t_k$ is the thickness of the substrate and kirigami layer, respectively. $E_s$ and $E_k$ is the approximated Young's modulus of the substrate and kirigami layer, respectively.  

While the z position of the neutral axis for the bilayer is calculated as,
\begin{equation}
    \begin{aligned}
        \bar{z}=\frac{t_k(t_k/2+t_s)+(E_s/E_k)t_s(t_s/2)}{t_k+(E_s/E_k)t_s}
    \end{aligned}
    \label{be2}
\end{equation}

For the bilayer structures, the equivalent bending stiffness is,
\begin{equation}
    \begin{aligned}
        D_\textrm{bi} = n\frac{{E_{k}[{t_k^3/12+t_k(t_k/2+t_s-\bar{z})^2}]+E_{s}[{t_s^3/12+t_s(t_s/2-\bar{z})^2}]}}{{\left( {1 - {\nu ^2}} \right)}}+(1-n)\frac{E_{s}{t_s^3/12}}{{\left( {1 - {\nu ^2}} \right)}}
        \end{aligned}
    \label{be3}
\end{equation}

\begin{table}
\centering
\caption{Material parameters for the substrate and Kirigami layer}
\begin{tabular}{llll}
\noalign{\smallskip} \hline \noalign{\smallskip}
Parameter & Value & Unit \\
\noalign{\smallskip} \hline \noalign{\smallskip}
Material constant $C_1^\textrm{s}$&  22.1 & KPa \\
Material constant $C_2^\textrm{s}$ & 1.7 & KPa\\
Thickness $t^\textrm{s}$ & 1.1 & mm\\
Material constant $C_1^\textrm{kg}$ & 17.9 & KPa \\
Material constant $C_2^\textrm{kg}$ & 84.5 & KPa\\
Thickness $t^\textrm{kg}$ & 1.6 & mm\\
Material constant $C_1^\textrm{kw}$ & -2.6 & KPa \\
Material constant $C_2^\textrm{kw}$ & 185.8 & KPa\\
Thickness $t^\textrm{kw}$ & 1.4 & mm\\

\noalign{\smallskip} \hline \noalign{\smallskip}
\label{table:matparam}
\end{tabular}
\end{table}








\medskip
\textbf{Supporting Information} \par 
Supporting Information is available from the Wiley Online Library or from the author.

\medskip
\textbf{Acknowledgements} \par 
We thank Shyan Shokrzadeh and Vishal Kackar for their assistance on experiments. Following research grants are gratefully acknowledged: NSF (CMMI-2053971) for L.M., M.M., and M.K.J.; and NSF (CAREER-2047663, CMMI-2101751, OAC-2209782) for M.K.J. 

\end{document}